\documentclass[a4paper]{article}

\pdfoutput=1

\usepackage[top=2cm, bottom=2cm, left=2cm, right=2cm]{geometry}
\usepackage{setspace}
\usepackage{epsfig}
\usepackage{amsfonts}
\usepackage{times}
\usepackage{graphicx}
\usepackage{subfigure}
\usepackage{epsfig}
\usepackage{url}
\usepackage{array}
\usepackage{amsfonts}
\usepackage{amsmath}
\usepackage{amssymb}
\usepackage{latexsym}
\usepackage{float}
\usepackage{multirow}
\usepackage{cite}
\usepackage{algorithm,algorithmic}
\usepackage{hyperref}
\usepackage{authblk}

\newcommand{\ben}{\begin{enumerate}}
\newcommand{\een}{\end{enumerate}}
\newcommand{\bi}{\begin{itemize}}
\newcommand{\ei}{\end{itemize}}
\newcommand{\be}{\begin{equation}}
\newcommand{\ee}{\end{equation}}
\newcommand{\bea}{\begin{eqnarray}}
\newcommand{\eea}{\end{eqnarray}}
\newcommand{\ba}{\begin{array}}
\newcommand{\ea}{\end{array}}
\newcommand{\bc}{\begin{center}}
\newcommand{\ec}{\end{center}}
\newcommand{\bt}{\begin{tabular}}
\newcommand{\et}{\end{tabular}}
\newcommand{\bfig}{\begin{figure}[htb]}
\newcommand{\efig}{\end{figure}}

\newcommand{\comments}[1]{}

\begin{document}
%
\title{Bayesian one-mode projection for dynamic bipartite graphs}


\author[1,2]{Ioannis Psorakis \thanks{ioannis.psorakis@eng.ox.ac.uk}}
\author[1]{Iead Rezek}
\author[1]{Zach Frankel}
\author[1]{Stephen J. Roberts}

\affil[1]{Pattern Analysis and Machine Learning Research Group, University of Oxford}
\affil[2]{Edward Grey Institute, University of Oxford}

\maketitle

\begin{abstract}

We propose a Bayesian methodology for one-mode projecting a bipartite network that is being observed across a series of discrete time steps. The resulting one mode network captures the uncertainty over the presence/absence of each link and provides a probability distribution over its possible weight values. Additionally, the incorporation of prior knowledge over previous states makes the resulting network less sensitive to noise and missing observations that usually take place during the data collection process. The methodology consists of computationally inexpensive update rules and is scalable to large problems, via an appropriate distributed implementation.



\end{abstract}

%

\section{Introduction}

A bipartite or one-mode network is a graph $\mathcal{G} = \{\mathcal{U,V,E}\}$ with two sets of nodes, $\mathcal{U}$ and $\mathcal{V}$, where connections $\mathcal{E}$ exist only between nodes that belong to different sets. The overall connectivity is described by the $N 
\times K$ \emph{incidence matrix} $\mathbf{B}$, where $N = |\mathcal{U}|$ and $K = |\mathcal{V}|$ and $b_{ik} = 1$ if there exists a link\footnote{although $b_{ik}$ may take any value in $\mathbb{R}$, denoting weight or participation strength, in this work we will consider only boolean incidence matrices.} between a given pair of nodes $i,k$ for which $i \in \mathcal{U}, k \in \mathcal{V}$ and zero otherwise. We use bipartite networks to describe a diverse range of complex systems; scientific collaboration networks \cite{newman_scientific_collaboration_networks_2}, animal visitation patterns to various sites \cite{psorakis_rsi,testing_association_patterns}, gene-to-disease associations \cite{human_disease_bipartite_network} Social Media \cite{konstas_collaborative_filtering}, product co-purchasing networks \cite{leskovec_viral_marketing}, and many more \cite{zhou_one_mode_projection,zweig_one_mode_projection}. 

One-mode projection is the operation where a bipartite network $\mathcal{G}  = \{\mathcal{U,V,E}\}$ described by the $N \times K$ incidence matrix $\mathbf{B}$ is mapped to a graph with only one class of nodes, $\mathcal{G}_U = \{\mathcal{U,E}_U\}$ via $\mathbf{B}:N \times K \rightarrow \mathbf{W}:N \times N$. The new connections are now placed between nodes of the set $\mathcal{U}$, which we shall call from now on the ``source'' set, based on the way they linked to nodes of the vanished ``target'' set. The most trivial way to build the adjacency matrix $\mathbf{W}$ of $\mathcal{G}_U$ would be to set $w_{ij} = 1$ if nodes $i,j \in \mathcal{U}$ have at least one common target $k$ in $\mathcal{G}$ and zero otherwise \cite{newman_scientific_collaboration_networks_1,barabasi_scientific_collaboration_networks}. A reasonable refinement  \cite{newman_book} involves setting the weight of each link as the total number common targets, or \emph{co-occurrences} $w_{ij} = \sum_{k=1}^K b_{ik}b_{jk}$ that $i$ and $j$ have across nodes in $\mathcal{V}$. In matrix terms, this is achieved by $\mathbf{W} = \mathbf{BB}^{\sf{T}}$.  Further extensions have been considered, such as moderating the weight by taking into account the exclusivity of co-occurrences \cite{newman_scientific_collaboration_networks_1} or introducing a saturation function \cite{saturation_function_one_mode_projection}, which moderate the projected link weight $w_{ij}$.



It is worth noting that the typical one-mode projection $\mathbf{W} = \mathbf{BB}^{\sf{T}}$ forces all nodes from $\mathcal{U}$ that point to a particular target node $k\in\mathcal{V}$ to form a fully connected subgraph. Thus each target node $k$ corresponds to a $d_k$-clique in $\mathcal{G}_U$, where $d_k = \sum_{i=1}^N w_{ik}$ the \emph{degree} of $k$. Due to the heavy-tail degree distribution on the target set \cite{degree_distribution_bipartite1}, there is a non-trivial number of nodes in $\mathcal{V}$ with such a high degree $d_k$ that make the projected network almost fully connected \cite{lambiotte_music}. Methods that can be employed to regulate such densification in the resulting graph $\mathcal{G}_U$, range from information filtering \cite{fortunato_information_filtering,tuminello_information_filtering} to defining appropriate null models that examine the statistical significance of the observed weights \cite{psorakis_rsi} or network motifs \cite{zweig_one_mode_projection}.

In the present work, we seek to one-mode project a temporal bipartite network $\mathcal{G}^{(t)} = \{\mathcal{U,V,E }^{(t)}\}, t\in \{1,...,T\}$ that is described by a sequence of incidence matrices $\{\mathbf{B}^{(t)}\}_{t=1}^T$. The one-mode projection at any given time point $t$ captures the associations between nodes $i,j \in \mathcal{U}$, by taking into account past and present link information from all steps $1$ to $t$. We require that all projected connections between nodes $i,j$ are appropriately weighted so that we take into account both the strength and the statistical significance of the association. Finally, we seek to model the uncertainty over the resulting topology, by placing probability distributions over the presence of each link. The model is formally presented in Section \ref{sec:bomp} and its application presented in Section \ref{sec:example}. In Section \ref{sec:discussion} we conclude with a short discussion and our roadmap for future applications and theoretical extensions.

\section{Bayesian one-mode projection}
\label{sec:bomp}

\subsection{Problem statement}
\label{sec:problem_statement}

Consider a setting where we observe a temporal bipartite network as a sequence of ``snapshots'' $\{\mathbf{B}^{(t)}\}_{t=1}^T$, where each incidence matrix $\mathbf{B}^{(t)}: N \times K$ describes the linkage of $N$ \emph{agents} to $K$ \emph{targets}\footnote{For the sake of simplicity, from now on we will assume that $N$ and $K$ are fixed for each $t$, although such constraint can be relaxed.}. Such dataset may describe the buying habits of $N$ customers where each month $t$ are performing purchases among a set of $K$ products, or the daily mobility patterns of $N$ Social Media users who perform check-ins at $K$ locations. Our key assumption is that there is an underlying \emph{association} or \emph{similarity} network $\mathbf{\Pi} \in \mathbb{R}^{N \times N}$ between agents, which directly affects the structure of $\{\mathbf{B}^{(t)}\}_{t=1}^T$ in the sense that very ``similar'' agents consistently point to the same targets and vice-versa. Our goal is to \emph{learn} the structure of $\mathbf{\Pi} \in \mathbb{R}^{N \times N}$ at every step $t$, by defining a Bayesian model that captures our belief over presence (or absence) of each link along with a probability distribution over its connection strength.

\subsection{Probabilistic model for graph links}
\label{sec:prob_model}


Given the observation sequence $\{\mathbf{B}^{(t)}\}_{t=1}^T$ described in Section \ref{sec:problem_statement}, let us isolate one particular timestamp $t$, so that $\mathbf{B} = \mathbf{B}^{(t)}$.  Each element $b_{ik}$ is 1 if agent $i$ links to target $k$ and zero otherwise while the sum $d_i = \sum_{k=1}^K b_{ik}$ is the total targets or \emph{out degree} of $i$. Let us now define an additional variable $x_{ij}$ that we will call \emph{opportunities}, which is the number of target nodes \emph{either} $i$ or $j$ link to; that is obtained by performing an element-by-element logical disjunction on the rows of  $\mathbf{B}$ and summing the elements of the resulting vector:

\begin{equation}
x_{ij} = \sum_{k=1}^K \mbox{OR}(b_{ik}, b_{jk}) \label{eq:opportunities}
\end{equation}

 A list of variables used in this paper is presented in Table \ref{tab:notation}.

\begin{table}[ht]
\caption{Notation}
\label{tab:notation}
\centering
\begin{tabular}{|c|l|}
\hline
Variable & Interpretation\\
\hline
\hline

$N$ & \# of source nodes.\\
$K$ & \# of target nodes. \\
$\mathbf{B}$ & $N \times K$ incidence matrix of bipartite graph. \\
$\mathbf{W}$ & $N \times N$ projection matrix. \\
$w_{ij}$ & \# of co-occurrences of agents $i$ and $j$. \\
$d_i$ & \# degree of agent $i$ based on $\mathbf{B}$. \\
$x_{ij}$ & \# of targets linked to by either \\ & \hspace{.2cm} $i$ or $j$ (opportunities)\\
$\pi_{ij}$ & $\in [0,1]$ attraction coefficient of $i,j$.\\
$\alpha_{ij}, \beta_{ij}$ & Beta distribution parameters. \\

\hline
\end{tabular}
\end{table}

Given the observed $N \times K$ incidence matrix $\mathbf{B}$, we begin by performing the standard weighted one-mode projection, getting the \emph{co-occurrence} matrix $\mathbf{W} = \mathbf{BB}^{\sf{T}}$. Each $w_{ij} = \sum_{k=1}^K b_{ik}b_{jk}$ represents integer-valued counts that we can model as a draw from a binomial distribution:

\begin{equation}
\label{eq:binom_sample}
w_{ij} \sim \mbox{Binom} (\pi_{ij} ; x_{ij}),
\end{equation}

with two parameters; the number of opportunities $x_{ij}$  and a \emph{bias term} $\pi_{ij} \in [0,1]$ that corresponds to our modelling assumption that there is a latent \emph{attraction coefficient} between all pairs $i,j$, which controls the extent to which opportunities $x_{ij}$ are manifested as co-occurrences $w_{ij}$ across targets. We view $\pi_{ij}$ as a measure of \emph{similarity} or \emph{association} between $i$ and $j$ and it is the key variable in our model; the one-mode projection we propose is a matrix $\mathbf{\Pi} \in \mathbb{R}^{N \times N}$ that contains all such $\pi_{ij}$. 

Based on Eq. (\ref{eq:binom_sample}), the probability of observing a particular number of co-occurrences, or link weight, $w_{ij}$ is given by:

\begin{equation}
\label{eq:binom_prob}
P(w_{ij} | \pi_{ij}, x_{ij}) = {x_{ij} \choose w_{ij}} \pi_{ij}^{w_{ij}} (1 - \pi_{ij})^{x_{ij} - w_{ij}},
\end{equation}

which is the \emph{likelihood} function of the observed weights $w_{ij}$. As our inference task is to describe the attraction coefficient $\pi_{ij}$ given the known $w_{ij}, x_{ij}$, a first approach would consist of maximising Eq. (\ref{eq:binom_prob}) w.r.t. $\pi_{ij}$. The trivial maximum likelihood (ML) solution to Eq. (\ref{eq:binom_prob}) yields $\hat{\pi}_{ij} = w_{ij} / x_{ij}$, which is deemed inconvenient for the following reasons:

\begin{itemize}
\item it makes our model sensitive to degenerate values of $w_{ij}$ and $x_{ij}$ that result from noisy observations of $\mathbf{B}$.
\item it provides a point estimate of $\pi_{ij}$, thus not capturing the \emph{uncertainty} on the attraction coefficient due to noise and missing observations.
\item it does not provide us a systematic framework for \emph{learning} $\pi_{ij}$, by exploiting both past and future observations of the bipartite network.
\end{itemize}

To overcome the above difficulties, we need to employ a Bayesian approach by working with the probability distribution over $\pi_{ij}$; we start with an initial $P(\pi_{ij})$ and \emph{revise}  at each step $t$ as we observe new values for $w_{ij}$ and $x_{ij}$. 

Recall that in Eq. (\ref{eq:binom_sample}) and (\ref{eq:binom_prob}) we have stated that the co-occurrences $i$ and $j$ depend on the opportunities and the attraction coefficient. This can be expressed via a graphical model in Fig. \ref{gmodel}, where such probabilistic dependencies are stated via arrows from nodes $x_{ij}$ and $\pi_{ij}$ pointing to $w_{ij}$. This allows us to express the probability of $\pi_{ij}$ as: 

\begin{figure}[h!]
\begin{center}
\includegraphics[angle=0,scale=0.55]{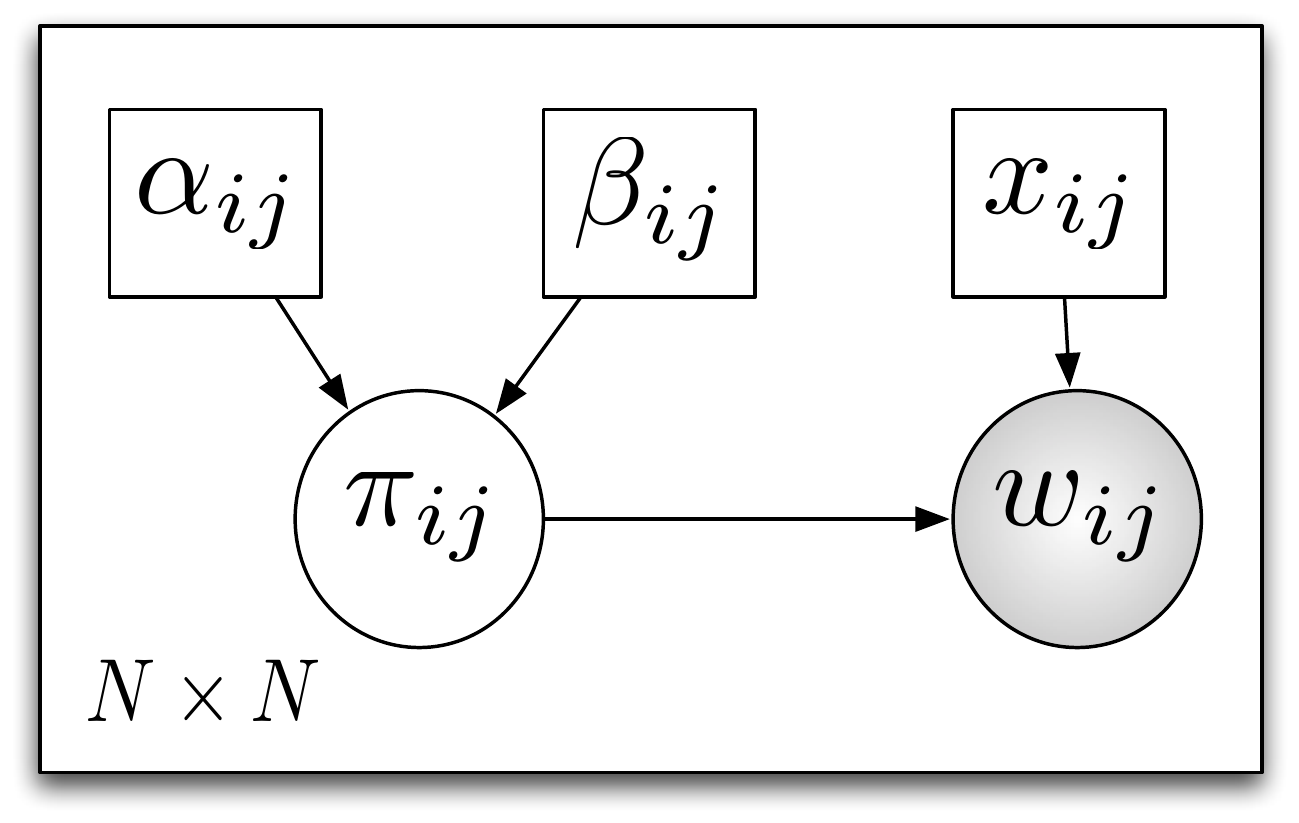}
\end{center}
\caption{\label{gmodel} Our graphical model, expressing how the observed (shaded circle) co-occurrences $w_{ij}$ between individuals $i$ and $j$ depend on the number of opportunities $x_{ij}$ (targets where either $i$ or $j$ link to in the original bipartite graph) and an unobserved (unshaded circle) attraction coefficient $\pi_{ij}$. The square plates denote deterministic parameters of the model.}
\end{figure}

\begin{equation}
\label{eq:posterior_p}
P(\pi_{ij} | w_{ij}, x_{ij}) = \frac{P(w_{ij} | \pi_{ij}, x_{ij}) P(\pi_{ij})} {\int_0^1 P(w_{ij},\pi_{ij} | x_{ij}) d\pi_{ij}},
\end{equation}

where $P(\pi_{ij})$ is the \emph{prior} and expresses our belief on how the attraction coefficient for $i,j$ varies before observing $w_{ij}$ and $x_{ij}$. On the other hand, the \emph{posterior} $P(\pi_{ij} | w_{ij}, x_{ij})$ is the revised belief on $\pi_{ij}$ under the light of these observations. Because $\pi_{ij} \in [0,1]$ we can model $P(\pi_{ij})$ as a Beta distribution:

\begin{equation}
\label{eq:beta_sample}
\pi_{ij} \sim \mbox{Beta} (\alpha_{ij},\beta_{ij}),
\end{equation}

parameterised by $\alpha_{ij}$ and $\beta_{ij}$, so that:

\begin{equation}
\label{eq:beta_prob}
P(\pi_{ij}) = \frac{ \pi_{ij}^{\alpha_{ij} - 1} (1 - \pi_{ij})^{\beta_{ij} - 1}} { \int_0^1 u^{\alpha_{ij} - 1} (1 - u)^{\beta_{ij} - 1} du}.
\end{equation}


Having an analytic structure for our prior in Eq. (\ref{eq:beta_prob}) , we combine it with the likelihood from Eq.  (\ref{eq:binom_prob}) based on Eq. (\ref{eq:posterior_p}) to get the posterior:

\small
\begin{eqnarray}
\label{eq:posterior_p}
P(\pi_{ij} | w_{ij}, x_{ij}) &=& \frac{P(w_{ij} | \pi_{ij}, x_{ij}) P(\pi_{ij})} {\int_0^1 P(w_{ij},\pi_{ij} | x_{ij}) d\pi_{ij}} \nonumber \\
                                &=&\frac{ {x_{ij} \choose w_{ij}} \pi_{ij}^{w_{ij}} (1 - \pi_{ij})^{x_{ij} - w_{ij}} } {\int_0^1 {x_{ij} \choose w_{ij}} \pi_{ij}^{w_{ij}} (1 - \pi_{ij})^{x_{ij} - w_{ij}}  d\pi_{ij}} \times \frac{ \pi_{ij}^{\alpha_{ij} - 1} (1 - \pi_{ij})^{\beta_{ij} - 1}} { \int_0^1 u^{\alpha_{ij} - 1} (1 - u)^{\beta_{ij} - 1} du} \nonumber \\
                                &=& \frac{\pi_{ij}^{w_{ij} + \alpha_{ij} -1 } (1 - \pi_{ij})^{x_{ij} - w_{ij} + \beta_{ij} -1}}{{x_{ij} \choose w_{ij}}^{-1} \times \int_0^1 {x_{ij} \choose w_{ij}} \pi_{ij}^{w_{ij}} (1 - \pi_{ij})^{x_{ij} - w_{ij}}  d\pi_{ij} \times \int_0^1 u^{\alpha_{ij} - 1} (1 - u)^{\beta_{ij} - 1} du} \nonumber \\
                                &=& \mbox{Beta} (\alpha_{ij}  + w_{ij}, \beta_{ij} + x_{ij} - w_{ij}) 
\end{eqnarray}
\normalsize

which is a revised Beta distribution over $\pi_{ij}$, with updated parameters:

\begin{eqnarray}
\alpha_{ij}' &=& \alpha_{ij} + w_{ij} \label{eq:update_a}\\
\beta_{ij}' &=& \beta_{ij} + x_{ij} - w_{ij} \label{eq:update_b}
\end{eqnarray}

The posterior distribution $P(\pi_{ij} | w_{ij}, x_{ij}) = \mbox{Beta} (\alpha'_{ij},\beta'_{ij}) $ provides all the information we need to describe the attraction coefficient $\pi_{ij}$, capturing the uncertainty over each possible value in $[0,1]$, while all dependencies between links are encoded in the $w_{ij}$ and $x_{ij}$ terms. In order to build the projection matrix $\Pi \in \mathbb{R}^{N \times N}$, fixed-point estimates can be directly derived from the posterior. For this particular study we have used the expected value $\mathbb{E}(\pi_{ij}) = \frac{\alpha_{ij}}{\alpha_{ij} + \beta_{ij}}$ for each element of  $\Pi \in \mathbb{R}^{N \times N}$.



Having from Eq. (\ref{eq:posterior_p}) a fully probabilistic formulation for the attraction coefficient, we can proceed one more step further and ``integrate out'' $\pi_{ij}$ from the likelihood function in Eq. (\ref{eq:binom_prob}) in order to obtain the probability distribution over the connection weight $w_{ij}$:  


\begin{eqnarray}
\label{eq:posterior_w}
P(w_{ij} | x_{ij}, \alpha_{ij}', \beta_{ij}') &=& \int_0^1 P(w_{ij}, \pi_{ij} | x_{ij}, \alpha_{ij}', \beta_{ij}') d\pi_{ij}  \nonumber \\
                                              &=& {x_{ij} \choose w_{ij}} B^{-1} (\alpha_{ij}',\beta_{ij}') \int_0^1 \pi_{ij}^{w_{ij} + \alpha_{ij}' -1} (1-\pi_{ij})^{x_{ij} - w_{ij} + \beta_{ij}' -1} d\pi_{ij} \nonumber\\
                                              &=& {x_{ij} \choose w_{ij}} \frac{B(w_{ij} + \alpha_{ij}', x_{ij} - w_{ij} + \beta_{ij}')}{B(\alpha_{ij}',\beta_{ij}')},
\end{eqnarray}

which is a Beta-binomial probability density function and $B(.,.)$ is the standard beta function. Such distribution captures the variability of co-occurrences $w_{ij}$ given our noise model. From the above equation we can estimate the expected value for the weights as $\mathbb{E}(w_{ij}) = \frac{x_{ij}\alpha_{ij}}{\alpha_{ij} + \beta_{ij}}$.

In this section we have described the theoretical foundation of our model along with the one-mode projection scheme for a single \emph{learning step} $t$. The full process involves cycling through the update equations:

\begin{eqnarray}
\alpha_{ij}^{(t)} &=& \alpha_{ij}^{(t-1)} + w_{ij}^{(t-1)} \label{eq:update_at}\\
\beta_{ij}^{(t)} &=& \beta_{ij}^{(t-1)} + x_{ij}^{(t-1)} - w_{ij}^{(t-1)} \label{eq:update_bt}
\end{eqnarray}

 and revising our distributions over the attraction coefficients and link weights. Details for the full learning scheme are presented in the following section.



\subsection{Algorithmic and implementation details}
\label{sec:algo_details}




Consider the state of the system at time $t=0$, before receiving the first network ``frame'' $\mathbf{B}^{(1)}$. At this stage, we have no observations regarding the bipartite graph and any prior beliefs on the agent pair associations $i,j$ are encoded in the Beta parameters $\alpha_{ij}^{(0)}, \beta_{ij}^{(0)}$. These can be initialised, for example, to vanilla values $\alpha_{ij}^{(0)} = \beta_{ij}^{(0)} = 10$ that center the attraction coefficients $\pi_{ij}^{(0)}$ around 0.5.

Upon receiving the first $\mathbf{B}^{(1)}$ we calculate the opportunities $x_{ij}$ and then the co-occurrences $w_{ij}^{(1)} = \sum_{k=1}^K b_{ik}^{(1)} b_{jk}^{(1)}$ for all $i,j$. We then update $\alpha_{ij},\beta_{ij}$ based on Eq. (\ref{eq:update_at}) and (\ref{eq:update_bt}). We then construct the projection matrix $\Pi$ using the expected values $\mathbb{E}(\pi_{ij})$. The whole process is presented in Algorithm 1.




\begin{algorithm}[ht!]
\label{algorithm}
\caption{Bayesian One-Mode Projection}
\begin{algorithmic}[1]
\REQUIRE bipartite sequence $\{\mathbf{B}^{(t)}\}_{t=1}^T$
\STATE Initialize $\alpha^{(0)}_{ij}, \beta^{(0)}_{ij}, \;\forall\; i,j \in \{1,...,N\}$

\FOR{$t = t_0$ to $T$}
\STATE Set $\mathbf{B} = \mathbf{B}^{(t)}$
\STATE Get opportunities $x_{ij}^{(t)}$ from Eq. (\ref{eq:opportunities})
\STATE Get co-occurrences via $\mathbf{W}^{(t)} = \mathbf{BB}^{\sf{T}}$

\FOR{$i,j \in \{1,...,N\}$}
\STATE update $\alpha_{ij}^{(t)}$ from Eq. (\ref{eq:update_a})
\STATE update $\beta_{ij}^{(t)}$ from Eq. (\ref{eq:update_b})
\STATE $\mathbb{E}^{(t)}(\pi_{ij}) = \frac{\alpha^{(t)}_{ij}}{\alpha^{(t)}_{ij} + \beta^{(t)}_{ij}} $
\STATE $\mathbb{E}^{(t)}(w_{ij}) = \frac{x^{(t)}_{ij}\alpha^{(t)}_{ij}}{\alpha^{(t)}_{ij} + \beta^{(t)}_{ij}}$
\ENDFOR

\ENDFOR

\RETURN $\mathbf{\Pi}^{(t)} = [\mathbb{E}^{(t)}(\pi_{ij})]_{i,j \in N}, \{\alpha_{ij}^{(t)}, \beta_{ij}^{(t)}, \mathbb{E}^{(t)}(w_{ij}), \forall\; t\in\{1,...,T\}$
\end{algorithmic}
\end{algorithm}

The computational cost of Algorithm 1 can be moderated via an appropriate distributed implementation. Non-holistic matrix operations such as the multiplication $\mathbf{BB}^{\sf{T}}$ can be parallelised (examples for Map-Reduce are shown in \cite{mining_massive_datasets}) while $\alpha_{ij},\beta_{ij}$ updates for each pair $i,j$ can be performed at different processing units. The benign computational scalability of the method lies on the structure of the probabilistic model itself; the conjugacy of our Beta prior in Eq. (\ref{eq:beta_prob}) with our Binomial likelihood function in Eq. (\ref{eq:binom_prob}) makes the posterior in Eq. (\ref{eq:posterior_p}) an updated Beta, thus no sampling (such as Markov Chain Monte Carlo) schemes need to be employed. In the next section we will describe application of the above in a working example, using an artificially generated dataset.

\section{An illustrative example}
\label{sec:example}

\begin{figure}[h!]
  \begin{center}
    \subfigure[]{\label{posteriors_12:p}\includegraphics[angle=0,scale=0.25]{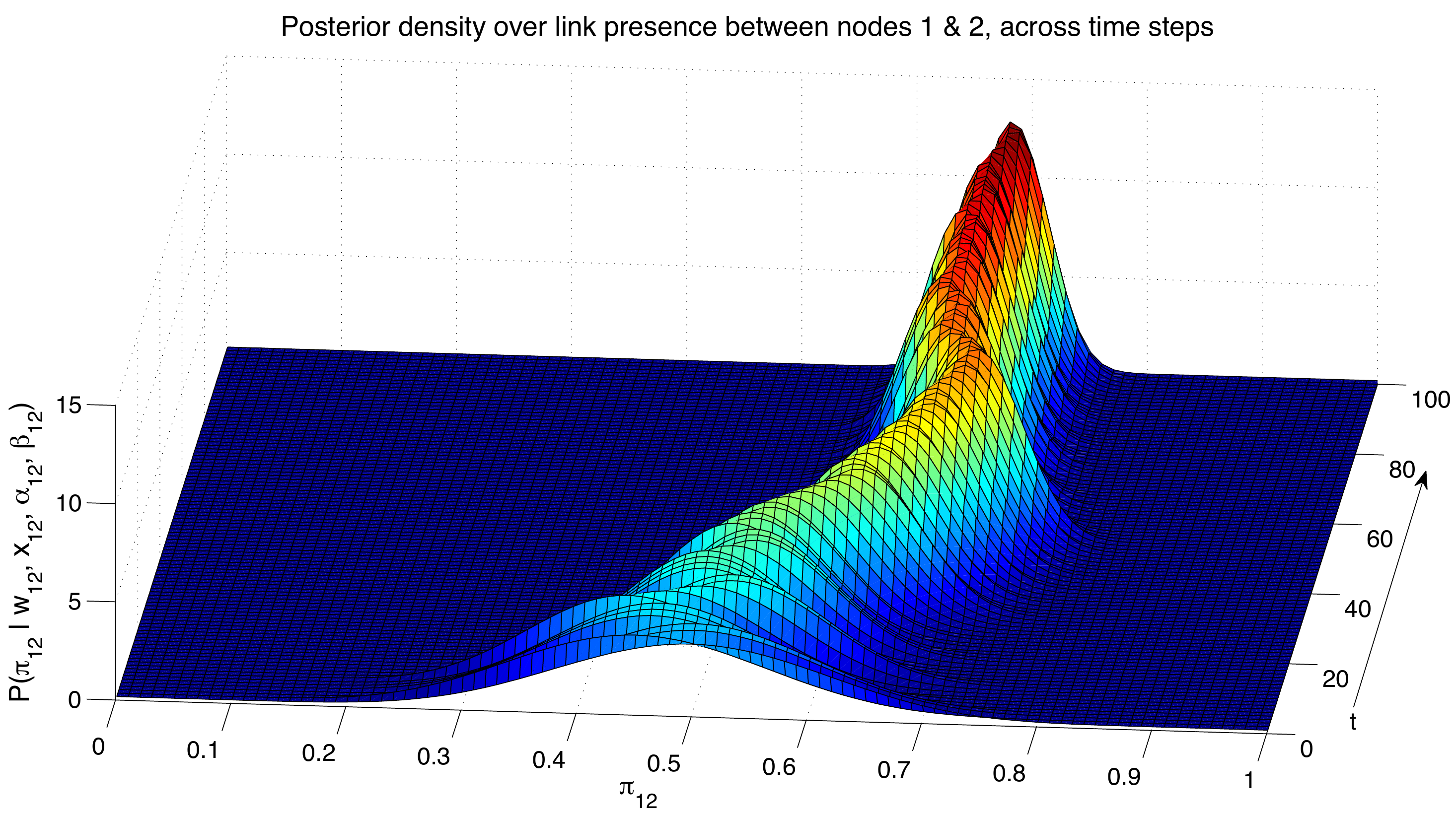}}
    \subfigure[]{\label{expectation_12:p}\includegraphics[angle=0,scale=0.25]{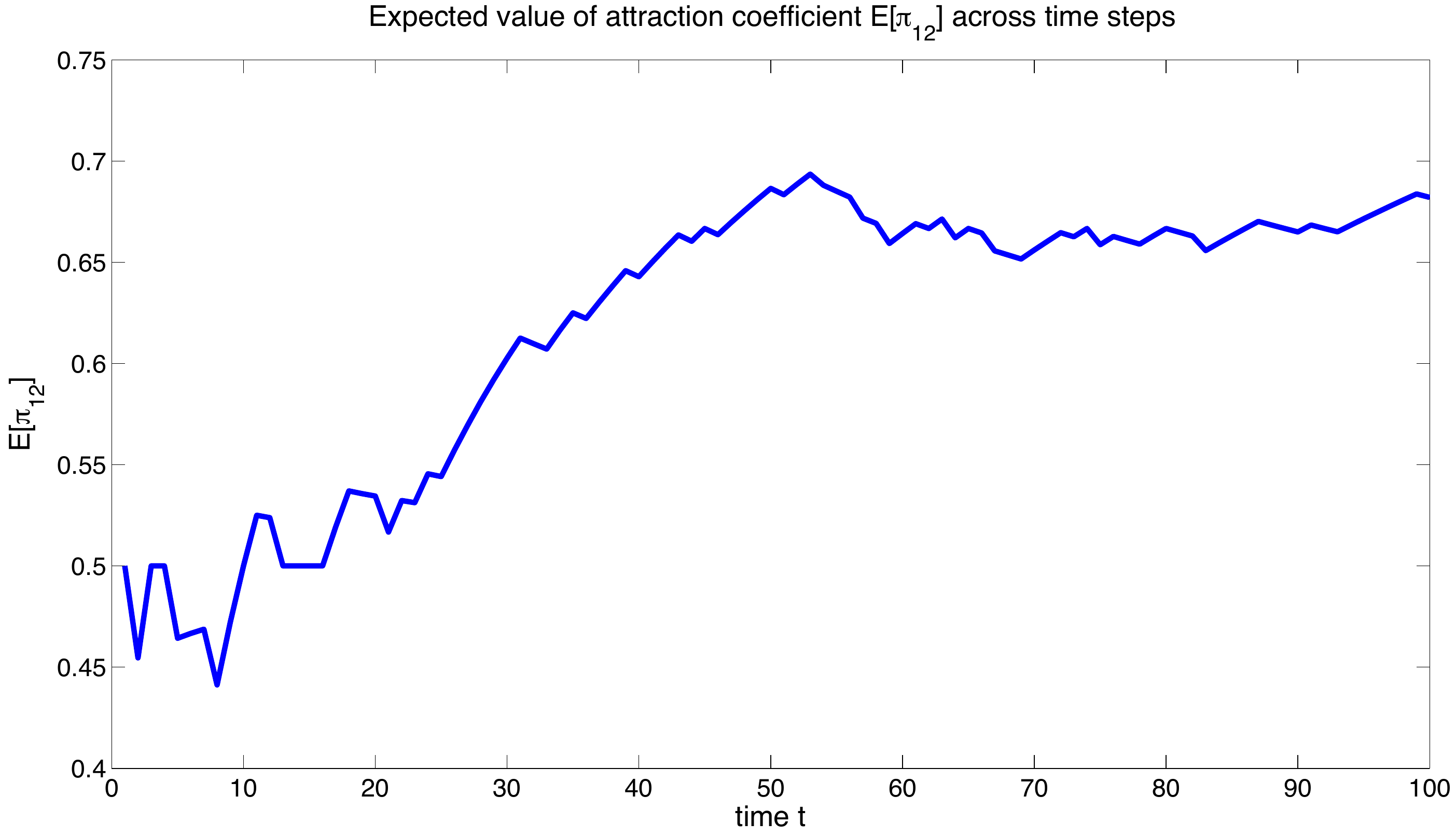}}
    \subfigure[]{\label{expectation_12:w}\includegraphics[angle=0,scale=0.25]{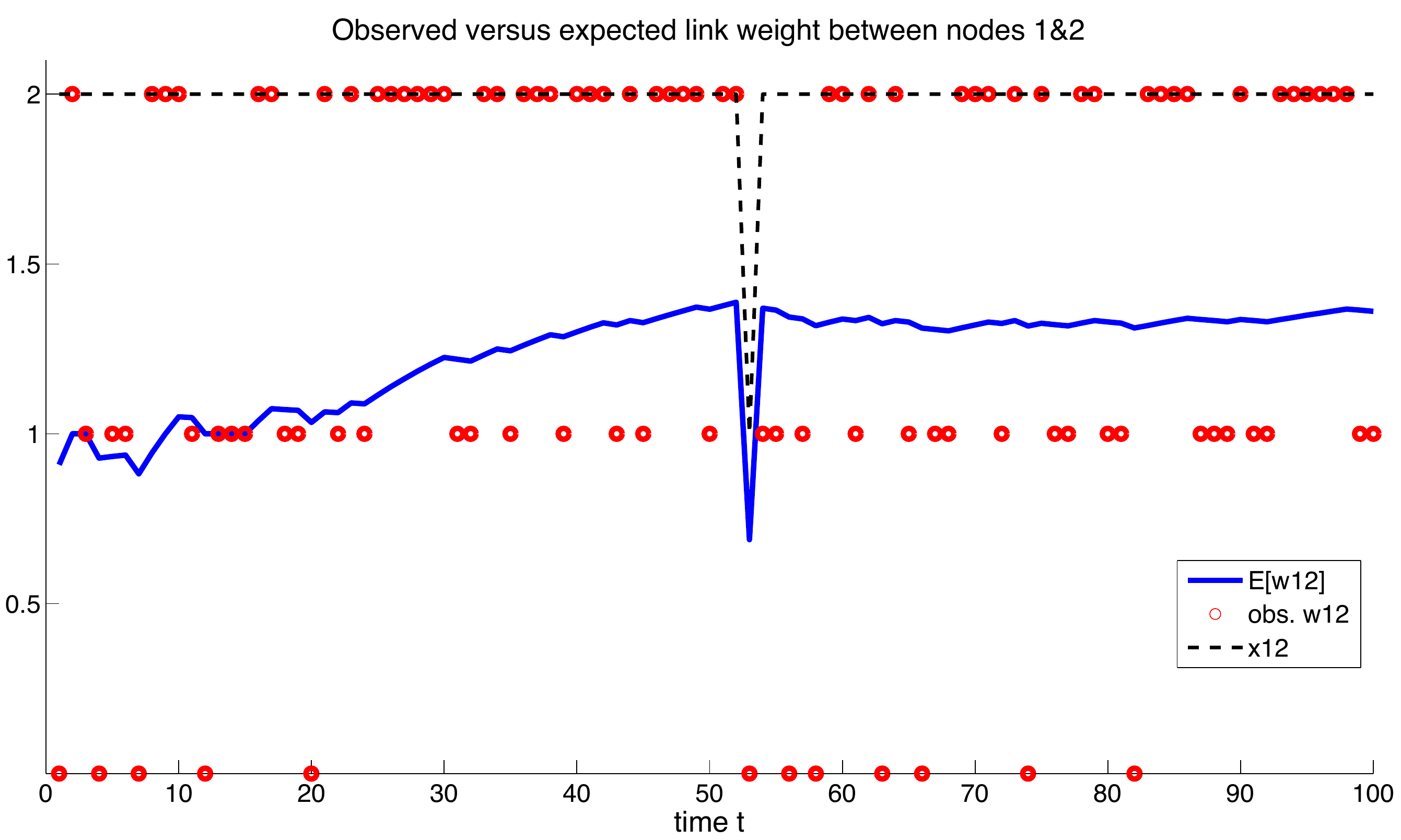}}
  \end{center}
  \caption{We plot various association metrics for the node pair 1-2, across 100 time steps. In Fig. \ref{posteriors_12:p} we illustrate how the distribution  over the attraction coefficient $\pi_{12}^{(t)}$ changes. Note that as we progress through time, not only the expected value increases, as shown in Fig. \ref{expectation_12:p}, but also probability mass is more concentrated around the mode, thus making more confident estimations due to accumulated data. In Fig. \ref{expectation_12:w} we show how by taking into account the distribution on $\pi_{12}$, we obtain smoother estimates (blue line) on the number of co-occurrences in contrast to the observed ones (red dots).}
\end{figure} 

\subsection{Artificial data generation scheme}
\label{sec:example_data_generation}

Consider the following $5 \times 4$ incidence matrix, which encodes the topology of a toy bipartite network:

\[
\mathbf{B} = \left[\begin{array}{cccc}
1 & 1 & 0 & 0 \\
1 & 1 & 0 & 0 \\
1 & 1 & 1 & 1 \\
0 & 0 & 1 & 1 \\
0 & 0 & 1 & 1 \end{array} \right]
\]

using the above matrix as a template, we generate a noise-contaminated sequence of bipartite networks $\{\mathbf{B}^{(t)}\}_{t=1}^T$ based on a seed matrix:

\[
\mathbf{B}^{\mbox{seed}} = \left[\begin{array}{cccc}
0.80 & 0.90 & 0 & 0 \\
0.90 & 0.70 & 0 & 0 \\
0.90 & 0.80 & 0.90 & 0.90 \\
0 & 0& 0.80 & 0.90 \\
0 & 0& 0.60 & 0.90 \end{array} \right]
\]

For each day $t = \{1,...,100\}$, we generate $\mathbf{B}^{(t)}$ by iterating through all pairs $i,k$ and setting $b_{ik}^{(t)} = 1$ with probability of success $b_{ik}^{\mbox{seed}}$. Such temporal bipartite stream may represent $N=5$ products that are allocated every day to $K=4$ shopping baskets, during a period of $T=100$ days. At the end of each day $t$, we receive a new basket log $\mathbf{B}^{(t)}$. Product pairs 1-2 and 3-4 tend to be co-purchased quite frequently (possibly because they complement each other for a particular task), while product 3 tends to appear across all baskets due to its general popularity (e.g. milk). For each day $t$, our goal is to \emph{learn} the associations between each pair $i,j$, by exploiting  current and past observations. 




\begin{figure}[h!]
  \begin{center}
    \subfigure[]{\label{posteriors_24:p}\includegraphics[angle=0,scale=0.25]{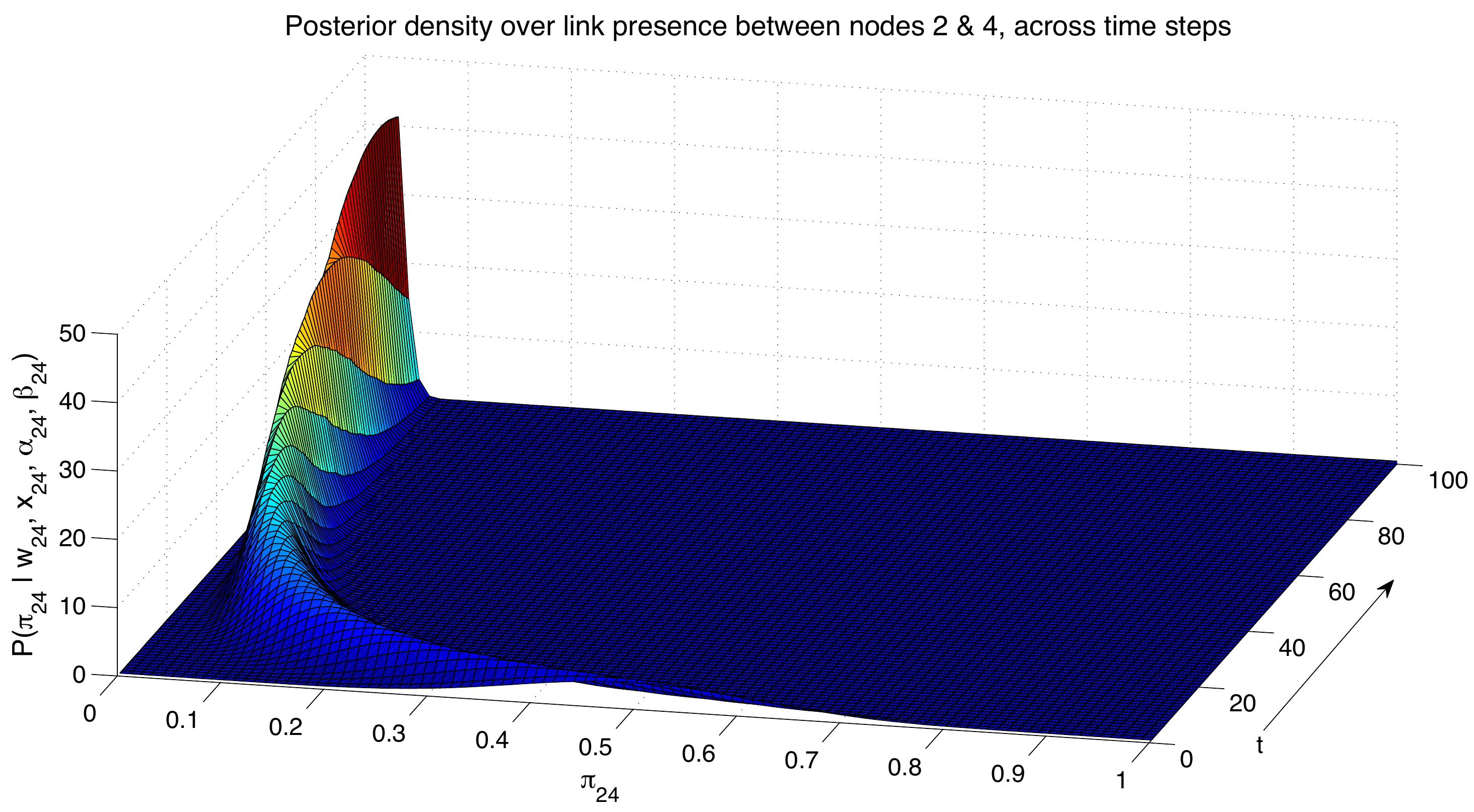}}
   \subfigure[]{\label{expectation_24:p}\includegraphics[angle=0,scale=0.25]{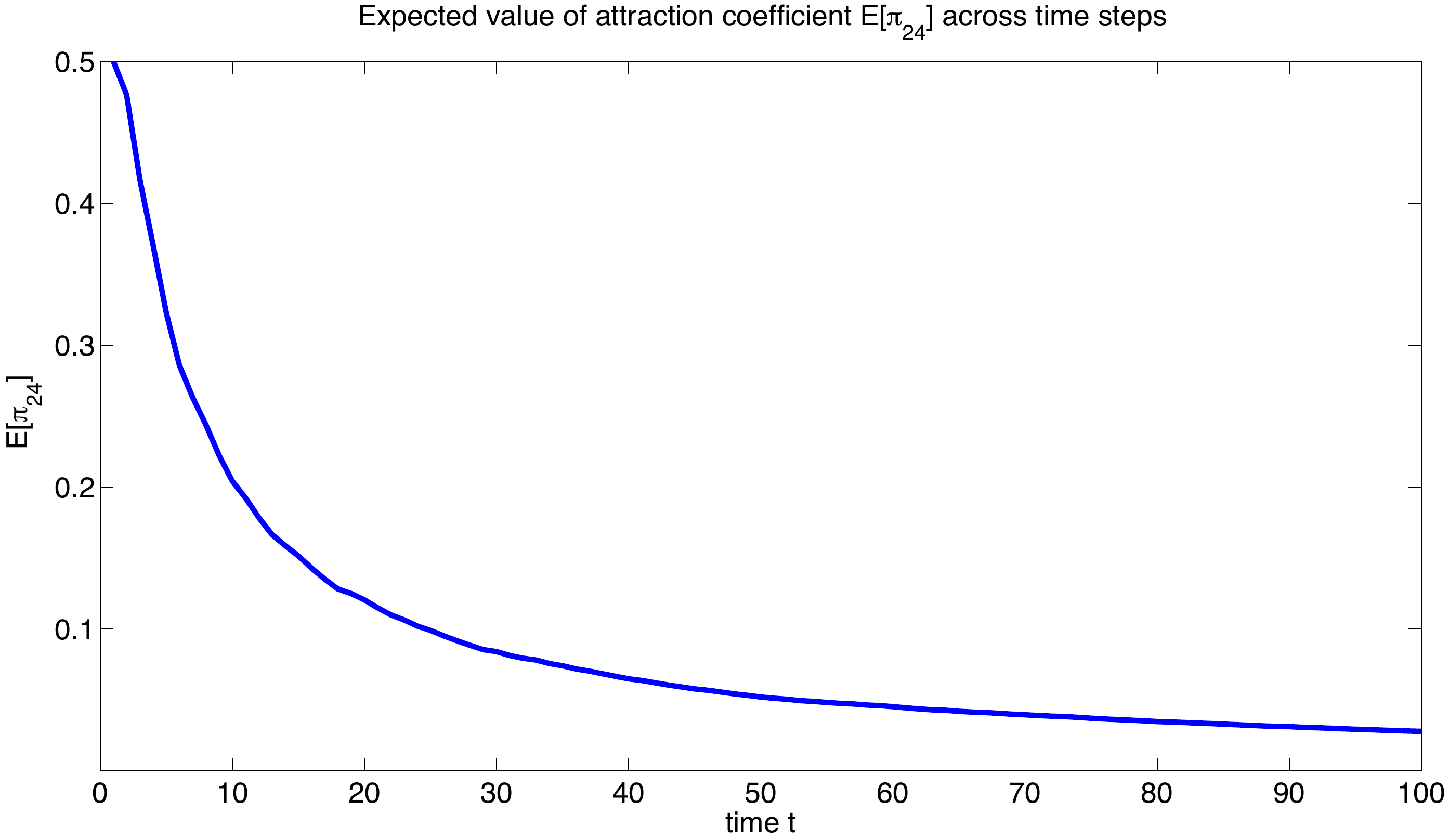}}
    \subfigure[]{\label{expectation_24:w}\includegraphics[angle=0,scale=0.25]{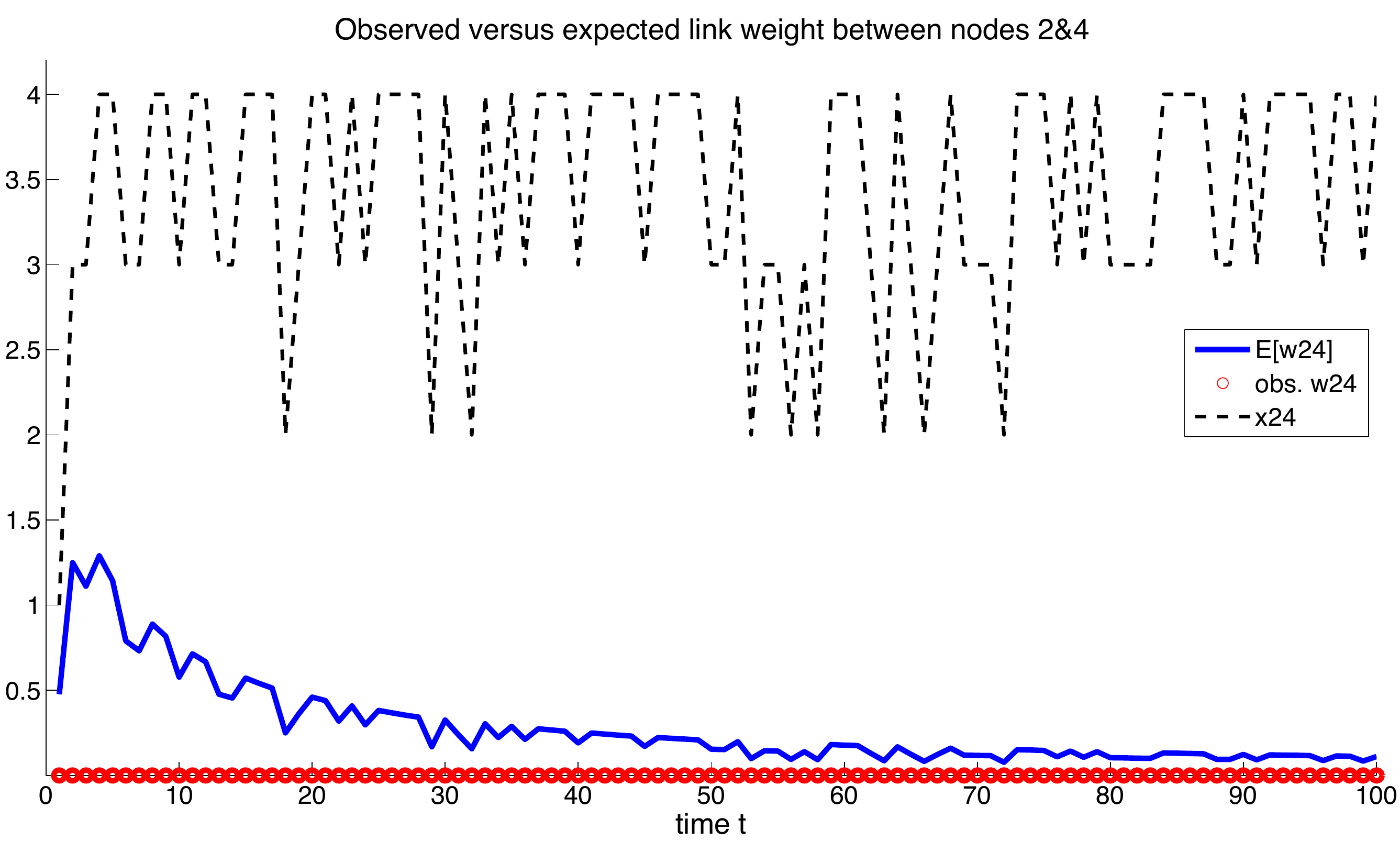}}
  \end{center}
  \caption{We plot various association metrics for the case of nodes 2-4, which do not have common targets in our sequence of bipartite graphs. Similar observations as in the case of 1-2, we get more confident predictions for $\pi_{24}$ in Fig. \ref{posteriors_24:p} and smooth estimates for their number of co-appearances in Fig. \ref{expectation_24:w}. The difference in this example is that our posterior belief on the attraction coefficient is dropping, as we do not observe any co-occurrences (red dot in Fig. \ref{expectation_24:w}) $w_{ij}$.}
\end{figure}

\subsection{Applying the method}

\begin{figure}[h!]
  \begin{center}
    \subfigure[]{\label{posteriors_23:p}\includegraphics[angle=0,scale=0.25]{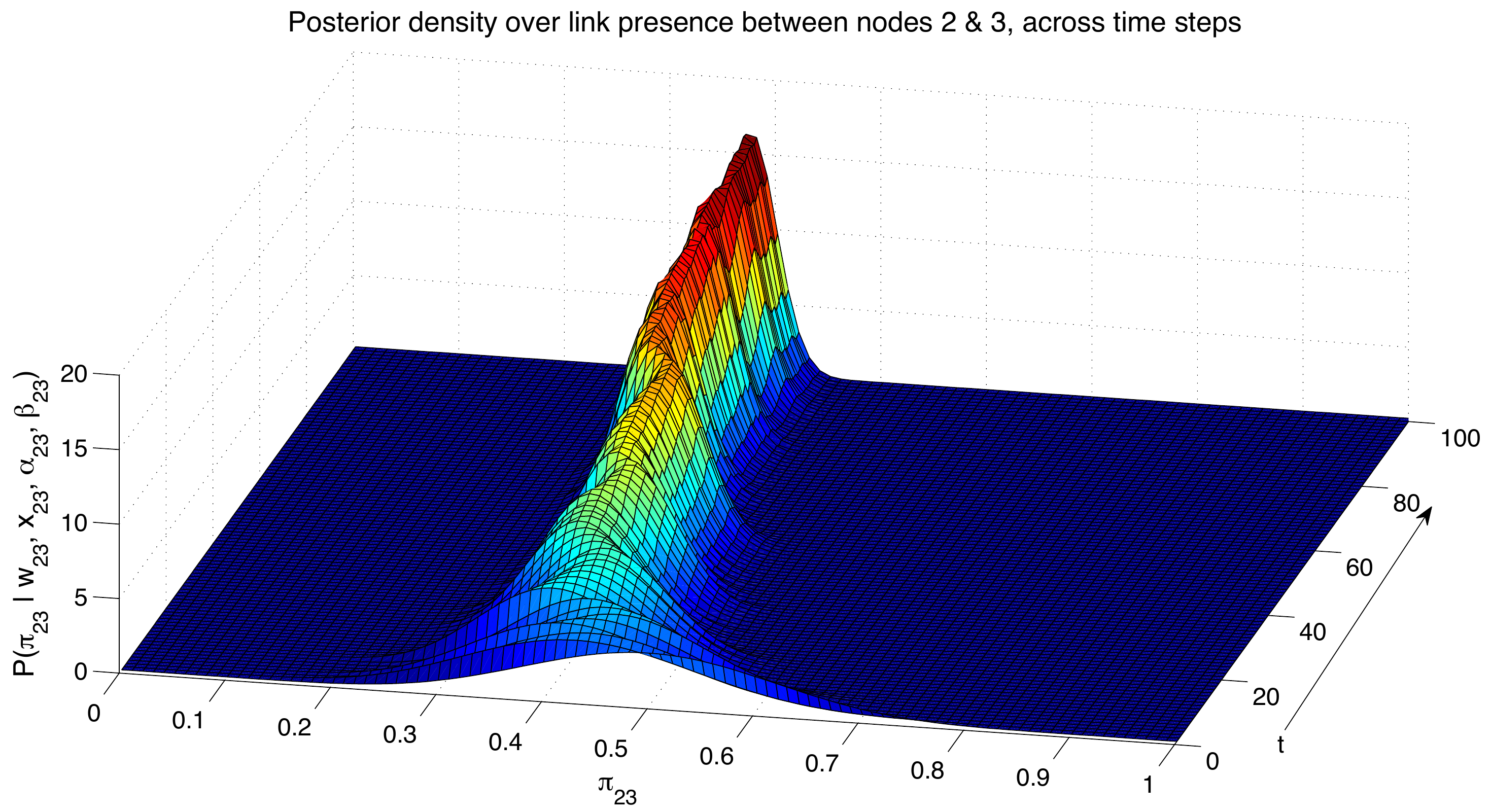}}
    \subfigure[]{\label{expectation_23:p}\includegraphics[angle=0,scale=0.25]{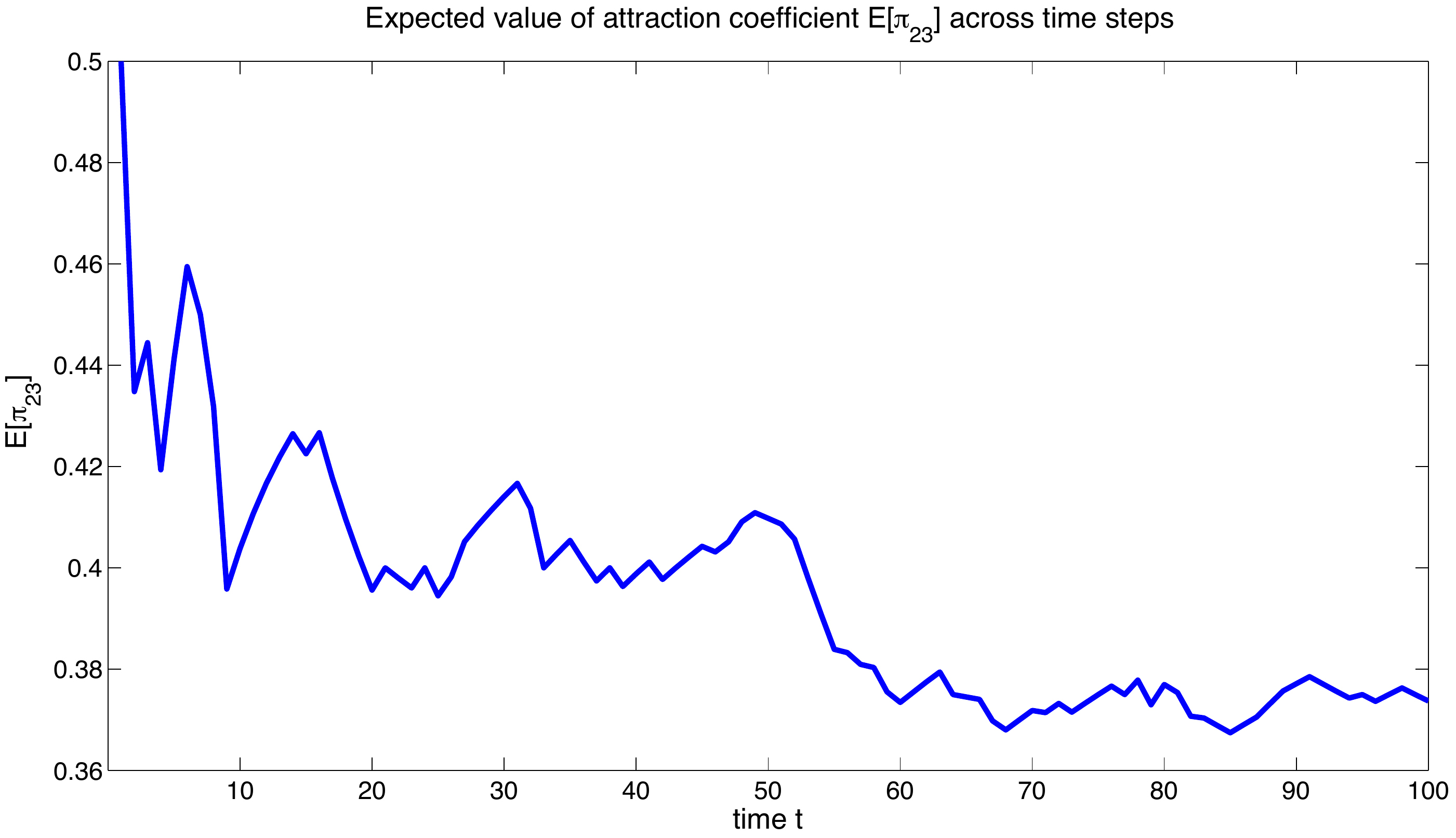}}
    \subfigure[]{\label{expectation_23:w}\includegraphics[angle=0,scale=0.25]{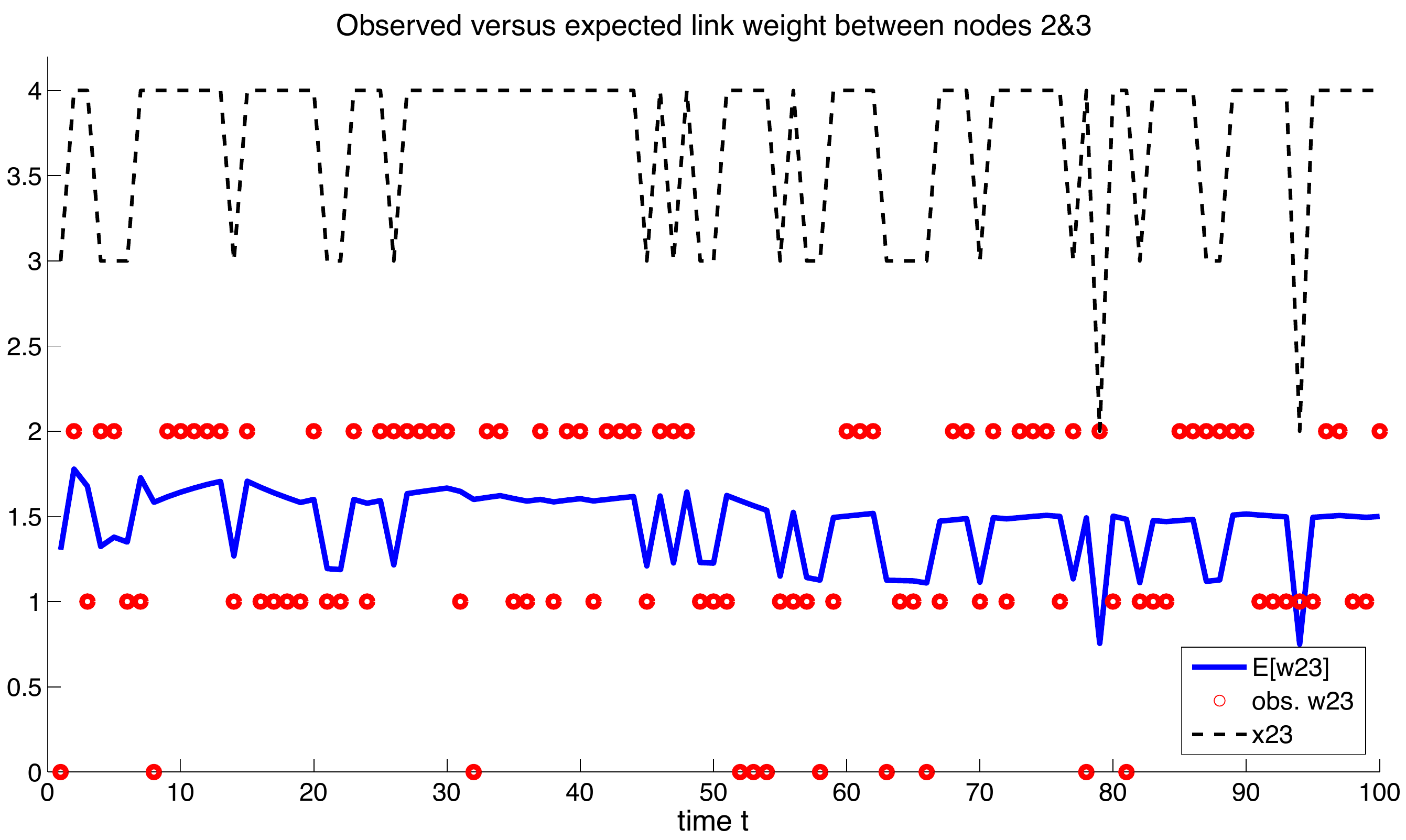}}
  \end{center}
  \caption{We plot association metrics for the node pair 2-3. Although the pair has about same number of co-occurrences across $t$ as 1-2 (red dots in Fig. \ref{expectation_23:w}), the estimated attraction coefficient decreases (see Fig. \ref{posteriors_23:p} and \ref{expectation_23:p}) due the large number of opportunities $x_{23}$ (black line in Fig. \ref{expectation_23:w}), that imply lack of exclusivity in the observed co-occurrences.}
\end{figure}


We start at $t=0$ by assuming no prior knowledge on any link structure. At this point, we have not seen any observations thus our model parameters must reflect our ignorance on how the associations may be. Therefore for each pair $i,j$ we set $\alpha_{ij}^{(0)} = \beta_{ij}^{(0)} = 10$ that gives $\mathbb{E}[ \pi_{ij}^{(0)} ]  = 0.5 \;\forall i,j$, denoting that we are unable to tell if there should be an association link  between $i,j$ before seeing any data.

At $t=1$ we have our first chunk of data, namely the incidence matrix $\mathbf{B}^{(1)}$. The opportunity values $x_{ij}^{(1)}$ are retrieved from $\mathbf{B}^{(1)}$ via Eq. (\ref{eq:opportunities}), while the co-occurrences $w_{ij}^{(1)}$ are retrieved via a standard weighted one-mode projection of $\mathbf{B}^{(1)}$ to $\mathbf{W}^{(1)}$. These values of $w_{ij}^{(1)}, x_{ij}^{(1)}$, along with $\alpha_{ij}^{(0)},\beta_{ij}^{(0)}$ from initialisation, are all that we need to describe the distribution of the attraction coefficients 
 via the update equations (\ref{eq:update_at}) and (\ref{eq:update_bt}). Once we have performed the update, we can get a fixed-value estimate of the attraction coefficient (or association index in our example) via its expected value under the posterior distribution $\mathbb{E}^{(t)}(\pi_{ij}) = \frac{\alpha^{(t)}_{ij}}{\alpha^{(t)}_{ij} + \beta^{(t)}_{ij}} $.

The posterior distribution over $w_{ij}^{(1)}$ (i.e. the number of baskets where both $i$ and $j$ co-appear) is obtained via Eq. (\ref{eq:posterior_p}), which gives a much better description of the  co-occurrences between $i$ and $j$, as it not only considers both the current and prior observations but also handles uncertainty in a principled manner and it is less sensitive to missing data.

In Fig. \ref{posteriors_12:p} we plot how the posterior distribution $P(\pi_{12} | w_{12}, x_{12}, \alpha_{12}, \beta_{12})$ of the attraction coefficient $\pi_{12}^{(t)}$ progresses during each iteration. We can see that for $t=0$ the distribution is our flat prior centered around 0.5, because we have no evidence to support the presence of an association between nodes 1 and 2. As we start observing non-zero link weights $w_{12}$ this prior belief is updated in order to explain the incoming data, effectively \emph{shifting} the distribution so that more probability mass is concentrated around larger values of $\pi_{12}$. Indeed, in Fig. \ref{expectation_12:p} we plot the expected value of $\pi_{12}^{(t)}$ for each $P(\pi_{12}^{(t)} | w_{12}^{(t)}, x_{12}^{(t)}, \alpha_{12}^{(t)}, \beta_{12}^{(t)})$ per iteration, where we can see the gradual increase of $\mathbb{E}[\pi_{12}^{(t)}]$ from $0.5$ to $\sim0.68$. It is important to notice that the increase of $\mathbb{E}[\pi_{12}^{(t)}]$ is less steep in later $t$, as the impact of new co-occurrences $i,j$ is not so strong as in the beginning of data collection. Such important saturation or  ``diminishing returns'' property arises naturally in Bayesian learning models, without the need to explicitly induce it via additional machinery such as hyperbolic tangent functions \cite{saturation_function_one_mode_projection}.  


On the other hand, in Fig. \ref{posteriors_24:p} we plot the posterior distribution $P(\pi_{24} | w_{24}, x_{24}, \alpha_{24}, \beta_{24})$ that expresses our belief on the presence of a link between nodes 2 and 4. As before, he distribution is initially centered around 0.5, due to the lack of evidence, while upon receiving data it rapidly shifts towards small values which can also be seen by plotting $\mathbb{E}[\pi_{24}]$ in Fig. \ref{expectation_24:p}. We also note that although the distribution is flatter during the initial iterations, we are constantly making more and more confident predictions (decreasing posterior entropy) as we keep observing $w_{24}^{(t)}$.

Another point worth noting, is the difference in behaviour of the observed link value $w_{ij}^{(t)}$ against the expected value $\mathbb{E}[w_{ij}^{(t)}]$ under the posterior distribution $P(w_{ij}^{(t)} | x_{ij}^{(t)}, \alpha_{ij}^{(t)}, \beta_{ij}^{(t)})$. In Fig. \ref{expectation_12:w} and \ref{expectation_24:w} we compare the observed $w_{ij}^{(t)}$ (red dots) against $\mathbb{E}[w_{ij}^{(t)}]$ where we can see that the expected value of the posterior is a \emph{smoother} estimate of the link weight, being less sensitive to large fluctuations due to noise and missing observations.

Now let us examine how our method models associations between $i=3$ and other nodes in the graph. Node 3 is an exceptional case in our example, as it tends to appear in every gathering target (recall the example where the milk tends to appear across all shopping baskets);  If we look again at the incidence matrix $\mathbf{B}^{\mbox{seed}}$ we can see that for every noisy realisation $\mathbf{B}^{(t)}$ of $\mathbf{B}$ in our artificial data stream, we have $P(b_{3k}^{(t)} = 1) \geq 0.8, \;\forall\; k,t$. Therefore under a naive model, node 3 would likely have a high similarity with all other nodes in the network, just because of its linkage to all targets in the original bipartite graph.

In Fig. \ref{posteriors_23:p} we plot the posterior density curves of $\pi_{23}^{(t)}$ across $t$, along with the expected values $\mathbb{E}[\pi_{23}^{(t)}]$ in Fig. \ref{expectation_23:p} and $\mathbb{E}[w_{23}^{(t)}]$ in Fig. \ref{expectation_23:w} as we did in the previous cases above. We can see that although we consistently observe non-zero link weights $w_{23}^{(t)}$ (red dots in Fig. \ref{expectation_23:w}), the association probability or attraction coefficient $\pi_{23}^{(t)}$ \emph{decreases} across $t$ (seen in Fig. \ref{expectation_23:p}). This is in complete disagreement with the case of association between agents 1 and 2, where consistently non-zero observations $w_{12}^{(t)}$ (red dots in  Fig. \ref{expectation_12:w}) led to an increase in the attraction coefficient $\pi_{12}^{(t)}$ (seen in Fig. \ref{expectation_12:p}). What is the reason behind this inconsistency?

The reason why our posterior belief over the presence of the link 2-3 is reduced, although we observe non-zero co-occurrences between individuals 2 and 3, is because of the role the opportunities variable $x_{ij}$ plays in the model. Recall Eq. (\ref{eq:binom_sample}):

\begin{equation}
w_{ij} \sim \mbox{Binom} (\pi_{ij} ; x_{ij}),
\end{equation}

which says that the observed link weight $w_{ij}$ is the result of a latent attraction term $\pi_{ij}$ that controls how many of the opportunities $x_{ij}$ are manifested as co-occurrences. These opportunities are the total number of targets where \emph{either} $i$ or $j$ link to. In the example of products 1 and 2 and based on $\mathbf{B}^{\mbox{seed}}$, their placement to baskets almost completely overlaps. Therefore, as we can see in Fig. \ref{expectation_12:w}, the observed number of co-occurrences $w_{12}$ (red dots) tends to match the number of opportunities $x_{ij}$ (black dashed line), forcing the model to infer high association values $\pi_{12}$.

On the other hand, although nodes 2 and 3 have similar values of $w_{23}$ with the observed link weights $w_{12}$ (red dots in Fig. \ref{expectation_23:w}), their number of opportunities $x_{23}$ (dashed black line in Fig. \ref{expectation_23:w}) is much higher due to the participation of agent 3 across all gathering targets. That leads our model, based on the binomial model in Eq. (\ref{eq:binom_sample}), to infer a lower value of the attraction coefficient $\pi_{23}$, effectively \emph{penalising such lack of exclusivity} in the co-appearances of 2 with 3. 

This is a very attractive property of the model, which not only regulates the link weights between ``gregarious'' node (that tend to link to everywhere) and ``selective'' ones (with a small set of targets they point to), but also allows the model to \emph{downplay the effect of purely coincidental co-appearances} on the attraction coefficient, which would otherwise introduce ``junk'' associations in the projection network.

\subsection{Notes on changepoint detection}
\label{sec:changepoints}

We have described a probabilistic model that allows us to infer latent associations between source nodes based on their common linkage to target nodes in a temporal bipartite network. 
Our approach consists of processing the data stream in a $T$-number of ``chunks'' and updating the model parameters based on new observations received at $t$ and prior knowledge from previous steps $0,...,t-1$. Such fusion of information from both current observations and past experience lies at the heart of every Bayesian learning model and allows us to perform rigorous inference in real-world settings where noise and missing observations are prevalent. Indeed, we have already shown in our artificial example of Fig. \ref{expectation_12:w} that although the observations (red dots) of the number of co-occurrences $w_{ij}$ between two agents greatly fluctuate, our probabilistic treatment allows us to extract a smooth trend of how $w_{ij}$ progresses through time.

We have shown that our method learns the association patterns of nodes, by making more and more confident predictions on the model quantities of interest while being resilient to perturbations induced by noise. The question is, what happens in cases where the underlying system dramatically changes at some given time point $t_c$, making all prior knowledge from $t=0,...,t_c -1$ obsolete?

In order to illustrate this, let us revisit the example of Section \ref{sec:example}, where at a given point $t_c$ we generate data from a different seed matrix:

%

%

\[
\mathbf{B}_c^{\mbox{seed}} = \left[\begin{array}{cccc}
0.90 & 0.80 & 0& 0\\
0& 0& 0.90 & 0.90 \\
0.80 & 0.90 & 0.80 & 0.90 \\
0.90 & 0.80 & 0& 0\\
0& 0& 0.90 & 0.80 \end{array} \right],
\]

where in this case products 1 \& 4 now co-appear in the first two baskets, 2 \& 5 in the last two while 3 remains common across all $K=4$ baskets.

\begin{figure}[h!]
\begin{center}
\includegraphics[angle=0,scale=0.25]{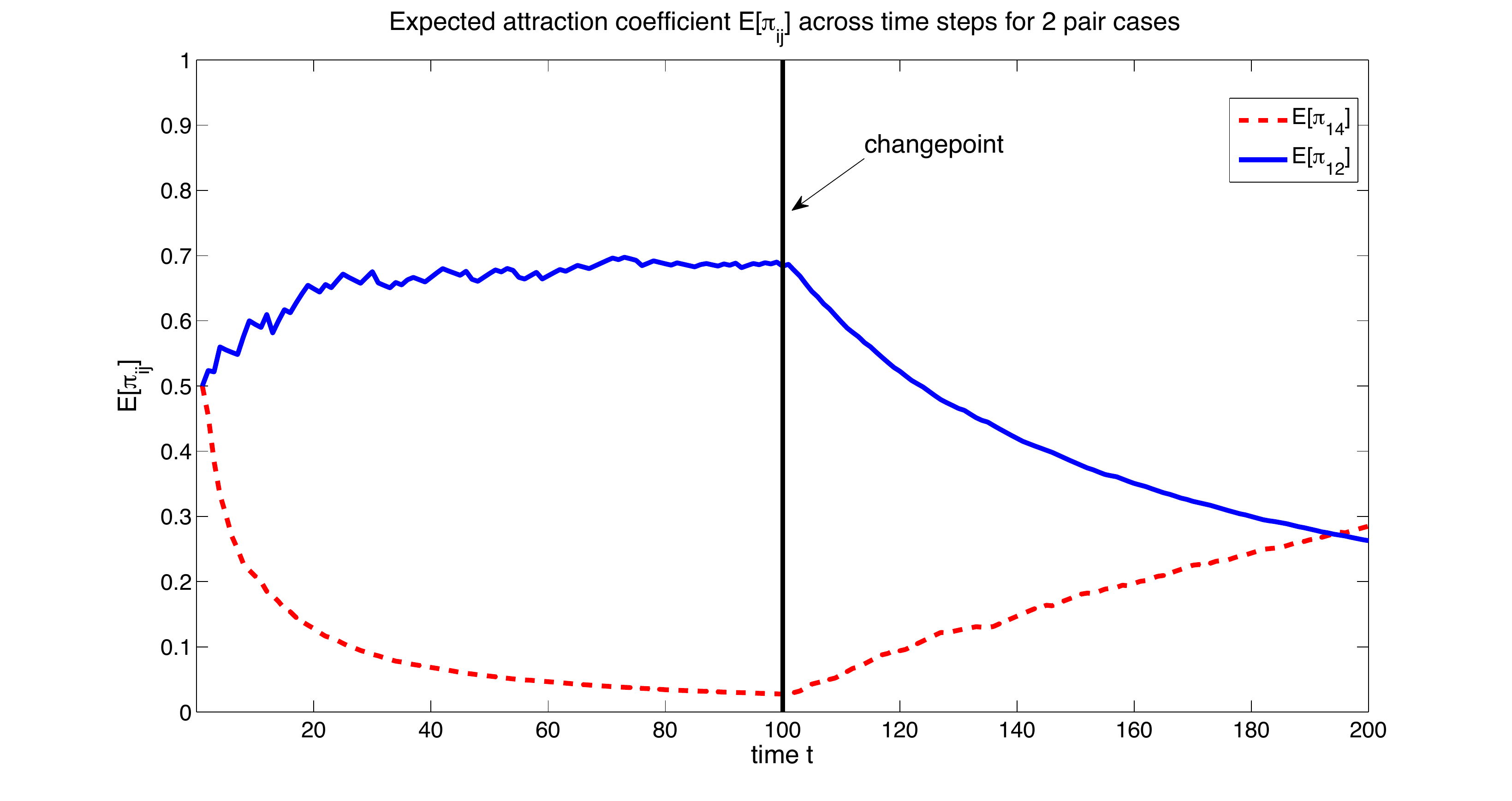}
\end{center}
\caption{\label{changepoint1} We demonstrate the responsiveness of the model at the presence of a particular changepoint. For the first 100 time steps, nodes 1 and 2 tend to point to the same targets in the bipartite network and model behaves exactly as in Fig. \ref{expectation_12:p}. After a particular step $t_c$, 1\&2 stop having common targets thus the attraction coefficient starts to drop. The reason for such slow drop after $t_c$ is due to the fact that past observations (before $t_c$) are strongly weighted in the model.}
\end{figure}

Assume now that we run our methodology on a dataset of $T=200$ instances of $\mathbf{B}^{(t)}$, where the first 100 are generated from $\mathbf{B}^{\mbox{seed}}$ (as in Section \ref{sec:example_data_generation}) and at $t_c = 101$ to $T$ we use the new $\mathbf{B}_c^{\mbox{seed}}$ instead. In Fig. \ref{changepoint1} we plot the expectation of the attraction coefficient $\mathbb{E}[\pi_{ij}]$ between pairs 1 \& 2 along and 1 \& 4.

We can see that for the first 100 steps, before the changepoint, the model has the exact behaviour as presented in Section \ref{sec:example}; as $w_{12}^{(t)}$ tends to be non-zero for $t<t_c$, the expected value of our posterior $\mathbb{E}[\pi_{12}^{(t)}]$ increases as we observe more data  (blue line in Fig. \ref{changepoint1}). Similarly, as $w_{24}^{(t)} = 0$ for $t<t_c$ there is a steep drop of $\mathbb{E}[\pi_{24}^{(t)}]$ to zero (dashed red line in Fig. \ref{changepoint1}). For $t>t_c$, we can see that the model responds by slowly shifting the posterior mean. In fact, even though the number of observations after the changepoint is the same as the one before $t_c$, the model fails to reach an appropriate value in both cases; there exists a lot of prior knowledge that makes the model to expect new data that relatively conform with the system state before $t_c$.

The limitation described above is not a drawback of Bayesian learning in general; in fact, the model behaves exactly as it should in settings where the underlying mechanism that generates the observations is \emph{stationary}. By implying such stationarity in the system, we are effectively making our model \emph{changepoint-naive} and our inferences very conservative as the status-quo is changing. In order to handle such cases, we have to control the way prior and current information is fused, by introducing a mixing coefficient $\kappa_{ij}$ that maps each update equation (\ref{eq:update_a},\ref{eq:update_b}) to a convex sum:

\begin{eqnarray}
\alpha_{ij}' &=& \kappa_{ij} \alpha_{ij} + (1-\kappa_{ij}  )w_{ij} \label{eq:update_ak}\\
\beta_{ij}' &=& \kappa_{ij} \beta_{ij} + (1-\kappa_{ij} )(x_{ij} - w_{ij}) \label{eq:update_bk}
\end{eqnarray}

At this stage, the parameter $\kappa$ is determined manually and depends on the application. To automate the selection of $\kappa$ the simplest approach is to use a bank of filters each with different values for $\kappa$. The choice of $\kappa$ can then be determined using the probability of the next observation $w_{ij}^{(t+1)}$ under the posterior predictive distribution \cite{Cesa_Bianchi} in Eq. \ref{eq:posterior_w}. 
While simple to implement, the approach could quickly become computationally expensive compared to alternative approaches based on generalised linear models which update the log-odds of the attraction coefficients using dynamic linear model recursions \cite{Roberts_Penny}. Both approaches avoid ad-hoc heuristics for the selection of $\kappa$ and allow us to detect change points and dynamically control the degree to which we mix past and present information in order to perform optimal predictions.


\section{Discussion and future work}
\label{sec:discussion}

In this report we have presented a probabilistic approach for one-mode projection temporal bipartite networks, in order to infer latent associations between the node class of interest. Such inferred associations  $\pi_{ij}$ are parameters of a binomial noise model, which can be viewed as link probabilities for all pairs of nodes, effectively mapping the the temporal bipartite network to an \emph{ensemble} of possible graphs. This is a very attractive aspect of our method, as, along with the distributions over $\pi_{ij}$, fully captures the uncertainty over connectivity patterns. Additionally, the model benefits for constant influx of new information by \emph{updating} our current beliefs over the network topology based on more recent observations. 

Capturing the uncertainty over graph links and weights is only one aspect of the method. The probability of connection or association between any pair of nodes an also be used for link completion tasks or personal recommendation tasks, while macroscopic topological properties of the inferred networks can now be described in the form of distributions over, for example, clustering coefficients, geodesic distances and diameters. That allows us to study the \emph{stability} of the overall structure and the crispness of properties such as community structure, homophily, small-world effect.


In addition to further development of theoretical methods, we plan to apply the one-mode projection to a variety of data domains well suited for the concept of temporal bipartite graphs, with highly dynamic and uncertain association structures. Namely, we have collected a dataset documenting the trading positions of many individual investors over time, containing both the date and the direction of their position. By treating specific investments within given time ranges  as the ``target events'' on our network - e.g. buying shares in Microsoft within two weeks of their earnings announcement, we develop a two-mode network of investors to events. 

We have already built a bipartite network of investors and investments, developing through this a non-temporal projection for an association network. That is to say, we trivially have the bipartite network of all stocks and traders if we do not consider time as a variable, and can project this onto the trader network to provide weighted associations. However, especially in finance, it is important to both capture the dynamics of associations as they change in time and quantify the uncertainty around such associations. Thus we have begun applying the method documented here to generate a probabilistic, and time dependent, network of associations among the traders. Given such network we can explore concepts such as homophily \cite{homophily}, assortativity \cite{newman_assortative_mixing} or community structure \cite{psorakis_nmf} among investors and their impact on returns. In future work we plan to run the model over a number of time periods to investigate the ability to detect changepoints, which in this case correspond to times of rapidly changing opinion or investing style (such as the financial crisis or the announcement of macro-economic policy shifts on the part of central banks). Preliminary results demonstrate the advantage of such an approach, but we plan to systematically explore its benefits, and compare it thoroughly to existing methods of detecting structure among investors.

Moreover, we plan to develop the method further to capture non-network related variables which vary over time, but impact the likelihood of association (in the case of financial markets, one can imagine general market sentiment, or stock correlation). The hope with this work would be to understand times at which joint appearance at target nodes represents stronger associations - if two investors make the same decision at times of limited liquidity, we should be able to capture this information systematically. In this sense, we regard the financial dataset as an ideal testing ground for future work as a variety of other variables are already tracked and easily experimented on. More generally, we hope that as we apply this method to further domains we will also discover deeper nuances which allow us to improve the theory underlying the model. 

\bibliographystyle{plain}
\bibliography{arXiv_BOMP}

\begin{thebibliography}{10}

\bibitem{barabasi_scientific_collaboration_networks}
A.L Barabási, H~Jeong, Z~Néda, E~Ravasz, A~Schubert, and T~Vicsek.
\newblock Evolution of the social network of scientific collaborations.
\newblock {\em Physica A: Statistical Mechanics and its Applications},
  311(3–4):590 -- 614, 2002.

\bibitem{Cesa_Bianchi}
N.~Cesa-Bianchi and G.~Lugosi.
\newblock {\em Prediction, learning, and games}.
\newblock Cambridge Univ Pr, 2006.

\bibitem{human_disease_bipartite_network}
Kwang-Il Goh, Michael~E. Cusick, David Valle, Barton Childs, Marc Vidal, and
  Albert-László Barabási.
\newblock The human disease network.
\newblock {\em Proceedings of the National Academy of Sciences},
  104(21):8685--8690, 2007.

\bibitem{konstas_collaborative_filtering}
Ioannis Konstas, Vassilios Stathopoulos, and Joemon~M. Jose.
\newblock On social networks and collaborative recommendation.
\newblock In {\em Proceedings of the 32nd international ACM SIGIR conference on
  Research and development in information retrieval}, SIGIR '09, pages
  195--202, New York, NY, USA, 2009. ACM.

\bibitem{lambiotte_music}
R.~Lambiotte and M.~Ausloos.
\newblock Uncovering collective listening habits and music genres in bipartite
  networks.
\newblock {\em Phys. Rev. E}, 72:066107, Dec 2005.

\bibitem{leskovec_viral_marketing}
Jure Leskovec, Lada~A. Adamic, and Bernardo~A. Huberman.
\newblock The dynamics of viral marketing.
\newblock {\em ACM Trans. Web}, 1(1), May 2007.

\bibitem{saturation_function_one_mode_projection}
Menghui Li, Ying Fan, Jiawei Chen, Liang Gao, Zengru Di, and Jinshan Wu.
\newblock Weighted networks of scientific communication: the measurement and
  topological role of weight.
\newblock {\em Physica A: Statistical Mechanics and its Applications},
  350(2–4):643 -- 656, 2005.

\bibitem{homophily}
M.~Mcpherson, L.~Smith Lovin, and J.~Cook.
\newblock {BIRDS OF A FEATHER: Homophily in Social Networks}.
\newblock {\em Annual Review of Sociology}, 27:415--445, 2001.

\bibitem{degree_distribution_bipartite1}
J.C. Nacher and T.~Akutsu.
\newblock On the degree distribution of projected networks mapped from
  bipartite networks.
\newblock {\em Physica A: Statistical Mechanics and its Applications},
  390(23–24):4636 -- 4651, 2011.

\bibitem{newman_scientific_collaboration_networks_2}
M.~E.~J. Newman.
\newblock Scientific collaboration networks. ii. shortest paths, weighted
  networks, and centrality.
\newblock {\em Phys. Rev. E}, 64:016132, Jun 2001.

\bibitem{newman_scientific_collaboration_networks_1}
M.~E.~J. Newman.
\newblock Scientific collaboration networks.$\hskip0.3em${}$\hskip0.3em${}i.
  network construction and fundamental results.
\newblock {\em Phys. Rev. E}, 64:016131, Jun 2001.

\bibitem{newman_assortative_mixing}
M.~E.~J. Newman.
\newblock Assortative mixing in networks.
\newblock {\em Phys. Rev. Lett.}, 89(20):208701, Oct 2002.

\bibitem{newman_book}
M.~E.~J. Newman.
\newblock {\em Networks: an Introduction}.
\newblock Oxford University Press, 2010.

\bibitem{Roberts_Penny}
W.D. Penny and S.J. Roberts.
\newblock Dynamic logistic regression.
\newblock In {\em Neural Networks, 1999. IJCNN '99. International Joint
  Conference on}, volume~3, pages 1562 --1567 vol.3, 1999.

\bibitem{psorakis_rsi}
I.~Psorakis, I.~Rezek, S.J. Roberts, and B.C. Sheldon.
\newblock Inferring social network structure in ecological systems from
  spatio-temporal data streams.
\newblock {\em J. R. Soc. Interface}, 9(76):3055--3066, 2012.

\bibitem{psorakis_nmf}
Ioannis Psorakis, Stephen Roberts, Mark Ebden, and Ben Sheldon.
\newblock Overlapping community detection using bayesian non-negative matrix
  factorization.
\newblock {\em Phys. Rev. E}, 83:066114, Jun 2011.

\bibitem{fortunato_information_filtering}
Filippo Radicchi, Jos\'e~J. Ramasco, and Santo Fortunato.
\newblock Information filtering in complex weighted networks.
\newblock {\em Phys. Rev. E}, 83:046101, Apr 2011.

\bibitem{mining_massive_datasets}
A.~Rajaraman and J.~Ullman~D.
\newblock {\em Mining of Massive Datasets}.
\newblock Cambridge University Press, 2011.

\bibitem{tuminello_information_filtering}
M.~Tumminello, T.~Aste, T.~Di~Matteo, and R.~N. Mantegna.
\newblock A tool for filtering information in complex systems.
\newblock {\em Proceedings of the National Academy of Sciences of the United
  States of America}, 102(30):10421--10426, 2005.

\bibitem{testing_association_patterns}
H.~Whitehead, L.~Bejder, and C.A. Ottensmeyer.
\newblock Testing association patterns: issues arising and extensions.
\newblock {\em Animal Behaviour}, 69(5):e1--e6, 2005.

\bibitem{zhou_one_mode_projection}
Tao Zhou, Jie Ren, Mat\'u\ifmmode \check{s}\else~\v{s}\fi{} Medo, and Yi-Cheng
  Zhang.
\newblock Bipartite network projection and personal recommendation.
\newblock {\em Phys. Rev. E}, 76:046115, Oct 2007.

\bibitem{zweig_one_mode_projection}
Katharina Zweig and Michael Kaufmann.
\newblock A systematic approach to the one-mode projection of bipartite graphs.
\newblock {\em Social Network Analysis and Mining}, 1:187--218, 2011.
\newblock 10.1007/s13278-011-0021-0.

\end{thebibliography}

\end{document}